# Optimized Mission Planning for Planetary Exploration Rovers

Alexander D. Lavin[1]

*Abstract*— The exploration of planetary surfaces is predominately unmanned, calling for a landing vehicle and an autonomous and/or teleoperated rover. Artificial intelligence and machine learning techniques can be leveraged for better mission planning. This paper describes the coordinated use of both global navigation and metaheuristic optimization algorithms to plan the safe, efficient missions. The aim is to determine the least-cost combination of a safe landing zone (LZ) and global path plan, where avoiding terrain hazards for the lander and rover minimizes cost. Computer vision methods were used to identify surface craters, mounds, and rocks as obstacles. Multiple search methods were investigated for the rover global path plan. Several combinatorial optimization algorithms were implemented to select the shortest distance path as the preferred mission plan. Simulations were run for a sample Google Lunar X Prize mission. The result of this study is an optimization scheme that path plans with the A* search method, and uses simulated annealing to select ideal LZ-path-goal combination for the mission. Simulation results show the methods are effective in minimizing the risk of hazards and increasing efficiency. This paper is specific to a lunar mission, but the resulting architecture may be applied to a large variety of planetary missions and rovers.

I. INTRODUCTION

The Google Lunar X Prize (GLXP) challenges a robotic spacecraft to capture and transmit to Earth a "Mooncast" of images and video before and after traveling 500 meters on the surface of the Moon. A team of collaborators at Carnegie Mellon University and Astrobotic Technologies, Inc. is developing a lunar rover in pursuit of the X Prize and associated milestones. The mission concept of operations (ConOps) calls for a lunar lander to touchdown and deploy a rover, which will then traverse at least 550 meters while exploring the area. The characteristics of this mission are typical of planetary surface exploration.

Previous studies have investigated the use of optimization algorithms for planetary exploration. Typically image-based, aerial and/or rover imagery is used with computer vision methods to identify rocks and craters as obstacles. Most approaches generate grid-based maps from this data with the objective of rovers avoiding obstacles [1], and some incorporate terrain mapping to grade the traversability as poor, low, moderate, or high [2], [3].

Another common feature in planetary rover path planning is the development of global planners to work in cooperation with the rover's autonomous and/or teleoperated guidance, navigation, and control (GNC). Typically the global path planning is implemented into a scheme to work offline to supplement the rover's online sensors. Standard control architecture in mobile robotics is a combination of local and global planners, organized as shown in Fig. 1. The reactive layer handles local information, with real-time constraints. The deliberative, or global, layer considers the entire world, likely requiring computation time proportional to the problem size [4]. This can be to verify rover's vision, and/or hazard-avoiding sensors [3].

Path planning in field robotics is dominated by search-based methods. The basic concept is a cellular grid map with goal and robot locations, and paths are searched throughout the grid to solve the problem [4]. Several search algorithms common in rover path planning include A*, D*, and Dijkstra's algorithm [2], [5]. These have been shown to efficiently plan rover routes, relatively recently implemented on the Mars Exploration Rovers and Mars Science Lab operated by NASA. The choice of algorithm, however, depends on the application. For example, focusing on working with teleoperation, China's Yutu rover used a simple, robust particle swarm optimization algorithm for global navigation points planning [6].

The methods presented here are similar to previous planetary exploration AI studies in that the aim is to create an efficient route plan to work with the mobile robot's local GNC. For the GLXP mission the rover will be teleoperated, with very little autonomy. The rover will sense its environment mainly from stereo vision, and the global path will serve as a guide of waypoints.

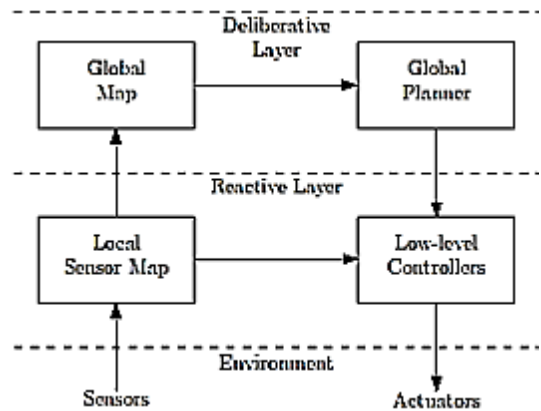

Figure 1. High-level block diagram of the standard hybrid control system architecture.

The novelty of the method presented in this study is integrating the optimization of both the landing zone (LZ) and rover path to select the most efficient LZ-path-goal combination. Proposed here is a combined scheme of a global optimization algorithm and search-based methods to plan the GLXP mission. The following sections detail the mission optimization scheme. Section II discusses setting up the optimization problem from raw imagery. Section III details the optimization scheme and algorithms. Test results are presented in Section IV, followed by discussion and conclusions in Section V.

---
[1] Master's student, Dept. of Mechanical Engineering, Carnegie Institute of Technology '14; alavin@alumni.cmu.edu

## II. OPTIMIZATION PROBLEM SETUP

The high-level ConOps calls for the lunar lander to touchdown in close proximity to the target destination – here the Lacus Mortis pit – deploy the rover, drive to the pit and explore, traversing at least 550m. The rover's start state is then the touchdown site, and the goal state is a location on the rim of the pit. The most efficient LZ-path-goal combination is that which minimizes the path distance (cost) deviation from 550m, while avoiding obstacles and terrain hazards. The optimization problem is to find the least-cost LZ-path-goal. Fig. 2 shows the high-level flow of the optimization scheme.

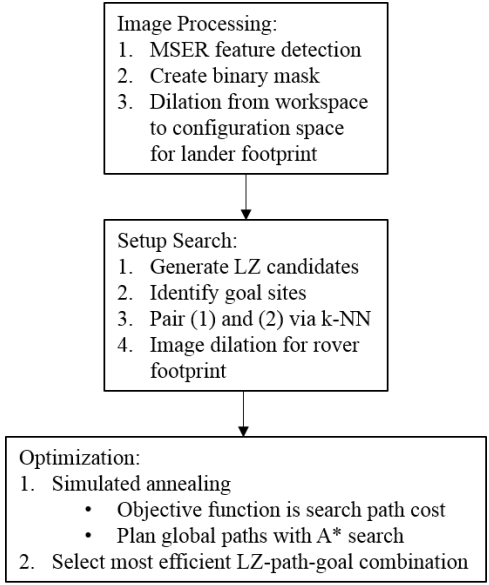

Figure 2. High-level flow diagram of the optimization scheme presented in this study.

### A. Image Processing and Obstacle Detection

Aerial imagery of the Lacus Mortis put site is sourced from the Lunar Reconnaissance Orbital (LROC) [7]. Craters, mounds, and rocks larger than two meters in width are classified as obstacles; two meters is chosen because it is large enough to obscure the landing, as well as cause the rover to deviate significantly from its path. Feature detection is used to identify objects in the images larger than this minimum size. Very large features represent terrain regions, e.g. a plateau or lake. Our interest lies with the smaller obstructions within the terrain regions. To identify these features in lunar imagery, the computer vision method of extracting maximally stable extremal regions (MSERs) is used. MSERs distinguish sets of distinct regions of a grayscale image, where the regions are defined by a threshold intensity across the region's outer boundary [8]. Consider the lunar image shown on the left of Fig. 3. The detected MSERs are stored as sets of pixels, highlighted red, pink, and blue in the right image of Fig. 3. It has been shown that extracting MSERs is of the most reliable object recognition methods, consistently outperforming other detectors in tests of viewpoint change, scale change, blurring, and light change [9]. These benefits are of interest because lunar imagery is inconsistent with regards to lighting conditions and shadows due to solar elevation changes, as well as off-nadir pointing.

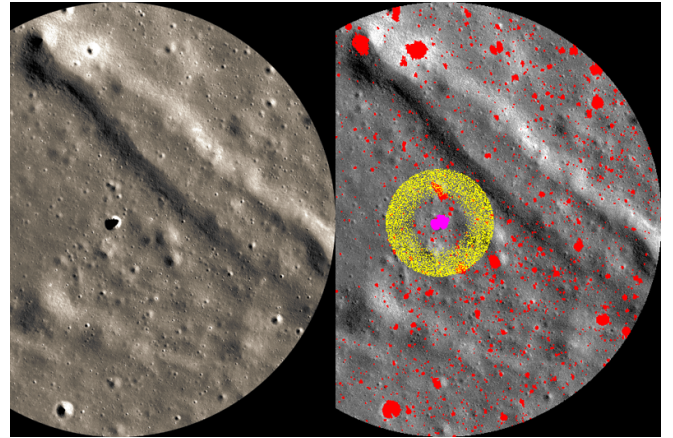

Figure 3. A lunar image from the dataset. Overlayed on the right image are extracted MSER obstacles (red), the Lacus Mortis pit (pink) and rim (blue), and generated LZ sites (yellow).

The locations on the rim of the Lacus Mortis pit, or the pixels on the border of the feature, are rover goal sites; these are colored blue in Fig. 1. With the GLXP competition requiring a minimum traverse of 550m, the preferred LZs are approximately this distance away from the goal sites on the pit rim. The set of lander goal sites are generated by approximating a band of points, with inner and outer radius from the pit center 250m and 750m, respectively. 10,000 candidate landing zone sites are generated, plotted in yellow overlaying the extracted obstacles in Fig. 3.

### B. Workspace, Configuration Space, and Solution Space

To translate the imagery into data useful for computation, image features are *masked* to create an occupancy grid, where free and occupied states are represented by 0s and 1s, respectively. This binary model is called the workspace $W$, where each of the free states make up the solution space $W_{free}$, and occupied cells make up $W_{obs}$. The set of landing zones is then the set of circular areas centered at each location in $W_{free}$. This is shown in the left diagram of Fig. 4.

Image dilation is used to transform $W$ into a configuration space $C$. This new space is calculated by eroding the edges around the free spaces, creating a morphology dilation of the *robot footprint*. For the lunar lander, the footprint is defined as a 50 meter diameter circle LZ. The transformation $W \rightarrow C$ narrows the solutions space: $C_{free} \subseteq W_{free}$. The dilation guarantees a LZ circle with a center in $C_{free}$ will not contain any obstacles [4], [10], [11].

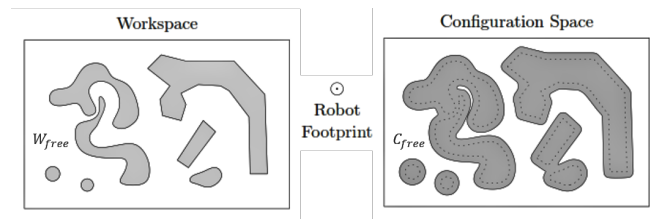

Figure 4. Dilation transformation from workspace $W$ (left) to configuration space $C$ (right). The white and shaded regions represent free space and obstacle space, respectively.

The result of these computations is the set of states in $C_{free}$ is the set of all possible landing zones, and subsequently the possible rover start sites. For the lunar image of Fig. 3, the dilation narrows the set of landing zones from 10,000 to 5,732 viable sites.

*B. Landing Zone and Path Planning Setup*

The lunar image is dilated as described above with a footprint to represent the LZ – i.e. a circle with 50m diameter. Each of the generated landing zones (yellow points in Fig. 3) falls in either $C_{free}$ or the rest of the configuration space, $C_{obstructed}$. That is, each location in the set of landing zones has a value of 0 or 1, representing free or obstructed, respectively. Only the states in the free space represent viable LZ locations.

The rover deploys from the lander deck vertically down to the surface, thus each viable landing zone is a start state for the rover's global path planning; a given LZ and start locations represent the same state. The size of the LZ guarantees a large obstruction-free area for the start of the rover path. For each start state, the K-nearest neighbors search method is used to find the nearest goal site on the pit rim (blue points in Fig. 3). Thus it is possible, and likely, that multiple start states share a common goal state. The resulting Euclidean distances approximate the closest start-goal pairs because the subsequent path plan may be shorter for a neighboring goal site.

The optimization objective is then to search for the most efficient path from the start states to corresponding goal states, minimizing path cost, which is calculated from the *costmap*. This is done by converting the occupancy grid to a costmap with a unit cost for traversing a free cell, and obstacle cells are represented by an infinite cost. The path cost, or total distance, is minimized for the global path plan of each start-goal state pair. The search methods used to generate the global path plans are discussed in Section III. The set of start-goal pairs make up the solution space.

## III. MISSION PLAN OPTIMIZATION

For each start-goal pair, an associated least-cost (minimum distance) global path plan can be calculated with search algorithms. The path of minimum cost in this set is determined to be the optimal start-path-goal combination for the mission. But calculating the comprehensive set of paths for all start-goal pairs (~6,000) is computationally expensive. Thus global optimization algorithms are used to find a start-path-goal combination that minimizes path cost.

*A. Global Optimization Algorithm*

The optimization problem is to find a good approximation of the global optimum in a large search space. A large solution space presents two priority concerns in optimization: (i) there can be many local minima, and (ii) an exhaustive search is computationally inefficient. And the problem is a *combinatorial optimization*, or one which seeks to find an optimal object from a finite set of objects [13]. That is, the solution space is the set of start-goal pairs, not a continuous function of values. Based on these problem parameters, *simulated annealing* is chosen as the optimization algorithm. The objective of simulated annealing is to locate the minimum cost configuration in the search space, where the objective function here is the rover path length for a given start-goal solution.

Simulated annealing is a global optimization algorithm, belonging to the families of metaheuristic and stochastic optimization methods. It is adapted from the Monte-Carlo algorithm, and designed for use with combinatorial optimization problems [14]. Inspired by the process of annealing in metallurgy, the method models the physical process of heating a material and then slowly lowering the temperature to decrease defects, thus minimizing the system energy; each solution in the search space represents a different internal energy of the system [15].

At each iteration a new configuration, or start-goal solution pair, is randomly pulled from the solution space. The distance of the new configuration from the current configuration, or the extent of the search, is based on a probability distribution with a scale proportional to the "temperature". The algorithm accepts all new points that decrease the objective function, but also, with a certain probability, points that increase the objective function. By accepting points that raise the objective, the simulated annealing algorithm avoids getting trapped in local minima in early iterations and is able to explore globally for better solutions [15]. "Slow cooling" is implemented in the simulated annealing algorithm as a slow decrease in the probability of accepting worse solutions as the algorithm explores the solution space. Accepting worse solutions is a fundamental property of metaheuristics, and beneficial for this problem because it enables a more extensive search for the optimal solution. As the solution space is explored over time, the probability of accepting worse solutions slowly decreases, or mimics slow cooling [16].

At each iteration of the simulated annealing algorithm the solution space is sampled, and a search method is then implemented to calculate the least-cost global path for the given start-goal pair; this is the subroutine in line 6 of the simulated annealing pseudocode of Alg. 1. The resulting path distance, or cost, is the "energy" function that simulated annealing uses to evaluate the start-path-goal configuration. The algorithm seeks for the configuration whose path cost minimizes the disparity from 550m.

Line 6 of the below pseudocode is a departure from the typical simulated annealing algorithm. It is specific to this problem in that the *current* value calculation requires a path plan algorithm to be nested within the energy evaluation.

**Algorithm 1** Simulated Annealing [17]
**Inputs**: *problem*, temperature *schedule*, stoppage criteria
**Output**: solution state
1  *current* ← *problem*.initial state
2  **for** i = 1 to max iterations
3    T ← *schedule*(i)
4    **if** T=0 then return *current*
5    *next* ← randomly selected successor of *current*
6    call subroutine pathSearch(*next*)
7    $\Delta E$ ← *next*.Value – *current*.Value
8    **if** $\Delta E$>0 **then** *current* ← *next*
9    **else** *current* ← *next* only with probability $e^{\Delta T/E}$
10 **return** *current*

The pathSearch() method uses the A* search algorithm, discussed in the next section. The temperature is *scheduled* to decrease exponentially, where $T_i = 0.95^{i-1} T_{i-1}$.

*B. Search Methods for Path Planning*

The search method, given start and goal states, aims to find the least-cost path between the states. This provides an evaluation metric for the simulated annealing optimization, and also a sequence of waypoints designating the most efficient rover path to follow. Before running search algorithms, a new configuration space $C$ must be generated with a 1 m-wide square morphological structure to represent a rover-sized footprint.

Search methods plan a path for the rover from start to goal states by expanding nearby states, or nodes, and checking which node yields the optimal step in the path. The rover is navigating a (relatively) known environment because $C$ is known before the mission execution. That is, there is *a priori* knowledge of the terrain and obstacles from lunar imagery. Thus uninformed search methods, like Dijkstra's breadth-first algorithm, can be ignored in favor of informed search methods. The general approach of these methods is best-first, in which the next node expanded is based on an *evaluation function*, $f(n)$: estimated cost of the cheapest solution through node $n$. The choice of $f(n)$ determines the search strategy. A bonus of informed search is including a *heuristic function* $h(n)$: estimated cost of the cheapest path from a state to the goal state. Greedy best-first search is built solely on this heuristic, where $f(n) = h(n)$, expanding the node closest to the goal. The A* algorithm is perhaps the most popular best search method, adding to the heuristic the cost to reach the node, $g(n)$. That is, $f(n) = h(n) + g(n)$. The search algorithm, looking for the cheapest path, tries (expands) the node with the lowest $f(n)$ [17], [18]. A* pseudocode is shown in Alg. 2.

To determine the optimal sequence of waypoints, the A* algorithm is a favorite for route search problems [3], [5]. For graph search, as opposed to tree search, a consistency condition is required to guarantee optimality. A heuristic is consistent if, for every node $n$ and every successor $n'$ of $n$ generated by any action $a$, the estimated cost of reaching the goal from $n$ is no greater than the step cost of getting to $n'$ plus the estimated cost of reaching the goal from $n'$:

$$h(n) \leq c(n, a, n') + h(n') \quad (1)$$

Norvig and Russel [17] explain clearly how the A* heuristic satisfies the consistency condition, and also that A* is *optimally efficient*: no other optimal algorithm is guaranteed to expand fewer nodes than A*. As long as a better-informed heuristic is not used, A* will find the solution at least as fast as any other method.

**Algorithm 2** A* Search
1  **Initialize** open and closed lists
2  Put the starting node in the open list
3  Define f, the cost function
4  **While** the open list is not empty
5      q ← node on open list with smallest f
6      Remove q from open list
7      Generate q's 8 successors, set their parents to q
8      **For** each successor
9          **If** successor is a goal, then stop search
10         successor.g ← q.g + distance between successor and q
11         successor.h ← distance from successor to goal
12         successor.f ← successor.g + successor.h
13         **If** a node with same position as successor is in the open list & has a lower f than successor, then skip this successor
14         **If** a node with same position as successor is in the closed list & has a lower f than successor, then skip this successor
15         **Else**, add the node to the open list
16     **End For**
17     Push q to the closed list
18 **End While**

The pitfall of A* in robot navigation is it is a static algorithm, which presents an issue when handling obstacles – i.e. encountering an obstacle in what was previously classified as a free space. When the configuration space changes by encountering a new obstacle, the current path is no longer valid. The location of the new obstacle is reclassified in the configuration space, and the A* algorithm must be re-run from the start. For this reason, the D* algorithm is commonly used in mobile robot navigation, including both Mars Opportunity and Spirit rovers. The D* algorithm is designed to be a dynamic version of A*, better-suited for execution in the field, yet yields suboptimal results [4], [19], [20]. The GLXP rover in its current developmental state uses global path planning offline, as a guide of waypoints for the teleoperators. Thus A* is the algorithm of choice for this mission planning application. Several solution paths are shown in Fig. 5.

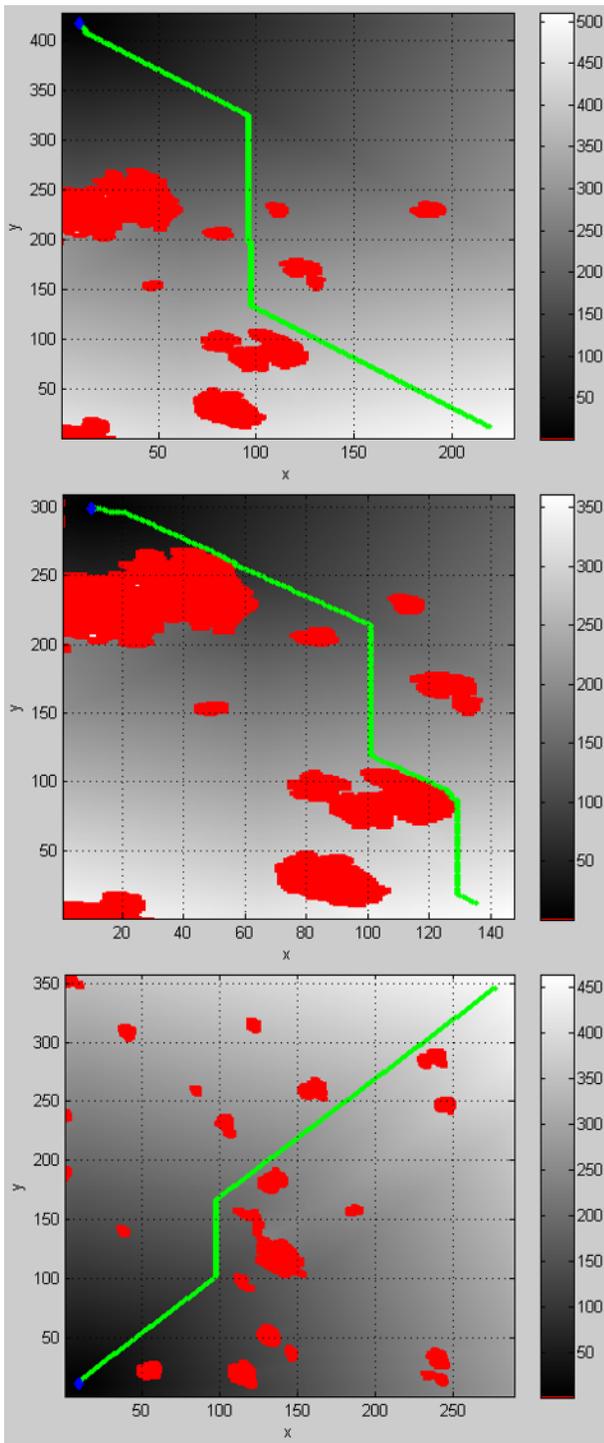

Figure 5.  Solution paths (green) avoiding red obstacles at three steps in the optimization scheme. The colorbar shows proximity to the goal state, which is marked with a blue diamond. From top to bottom, the solutions are at simulated annealing iterations 1 (initial), 10, and 100 (final). The initial solution is well over 550m, and the 10th solution is well under. The final solution is a global path plan of distance 541.816m. The path costs (function values) are plotted in Fig. 6.

With the simulated annealing global optimization algorithm selecting from a solution space of least-cost path plans, the mission optimization scheme uses the most efficient paths to select an optimal start-goal configuration. The result is an optimal combination of LZ and global path plan.

## IV.  RESULTS

The optimization scheme is run on a dataset of 15 lunar images of the Lacus Mortis pit, with varying lighting conditions and viewing angles. The images cover a circular region of 2500m diameter about the Lacus Mortis site. The detection of MSER features in the images is visually inspected in order to set the threshold and feature-size bounds at appropriate levels; visual inspection is satisfactory here because the dataset is relatively small. Simulated annealing is customizable to specific problems. Trials were run to tune the parameters for the temperature initial value and schedule function, and for termination criteria. The optimization scheme was written in MATLAB and run 45 times, three times per image, with results shown in Table 1.

TABLE I: OPTIMIZATION SCHEME TEST RESULTS

| Solution Path[a] | Average Cost | 9.2m |
|---|---|---|
| | Minimum Cost | 1.2m |
| | Maximum Cost | 14.4m |
| Unique Start-Goals | | 33 of 45[b] |
| Unique Paths | | 44 of 45[c] |

a. Solution path metrics are deviations from the objective 550m.
b. 12 of the total 45 start-goal configurations were repeats, where two were repeated twice.
c. One path (of the 12 possible repeated start-goal configurations) was repeated.

Differences in the optimal solution between trials of the same lunar image can be attributed to the simulated annealing algorithm settling on a solution that, although not the global optimum, is satisfactorily efficient. The simulated annealing process can be visualized with help of Fig. 6. Setting the iteration maximum at 100 was sufficient for all test trials. The objective (energy) function occasionally increases, yet significantly decreases overall. Differences in the optimal solution between trials of different lunar images, however, don't point directly to the optimization algorithm. There are small differences in the identified terrain obstacles, attributed to the varying lighting conditions and angles of view, influencing MSER feature detection with different shadows. Also the sets of LZ and rover goal states are inconsistent between images because of variations in the image resolutions.

The final output from the optimization scheme is a configuration of an LZ, global path plan, and goal destination. It is important to note, however, the resulting minimal-cost path is to serve as a guide for the rover operators, as opposed to an absolute route. The cell representation of the lunar terrain doesn't translate directly to rover driving, which may traverse at intermediate angles and curves. Also the rover may need to maneuver around previously unknown obstacles, or explore new areas of interest.

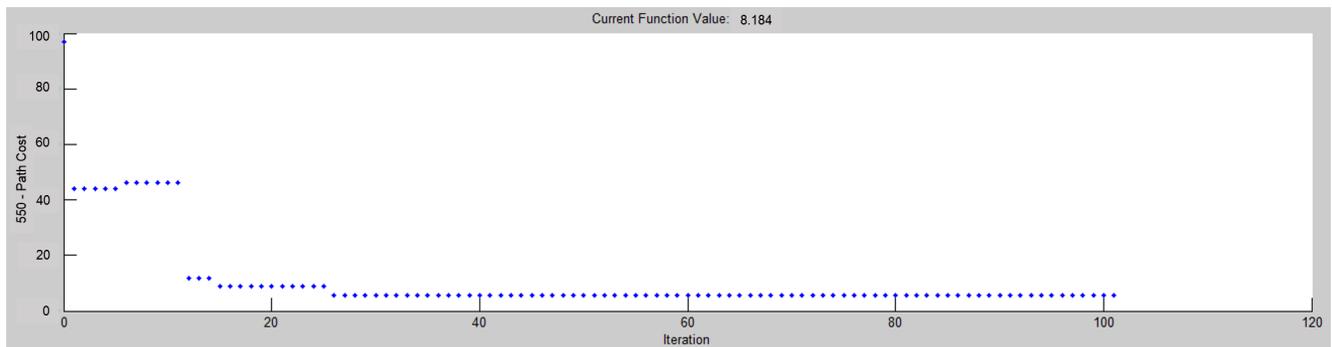

Figure 6. The energy function values at each iteration for one of the test trials – same trial as the path plots in Fig. 5. Here convergence was achieved at iteration 26, and 100 iterations was sufficient for all test trials.

## V. Discussion & Conclusion

This study presents a mission optimization scheme to contribute to future space exploration with planetary rovers. The simulated annealing and global path planning algorithms discussed in this paper attempt to address the need for efficient exploration methods.

Hazard identification is accomplished with two fundamental robotic vision methods: MSER feature extraction and image dilation. A combination search scheme is used to select the landing zone, rover path, and goal site for the mission. The rover path planning utilizes the A* search algorithm, providing a global path from landing zone to the Lacus Mortis pit, efficiently minimizing the path cost while avoiding obstacles. The path cost serves as the objective function in a simulated annealing optimization algorithm. Pairing the path planner with landing zone selection can significantly enhance the probability of mission success. And rover GNC benefits greatly by utilizing a global path planner to specify an optimal path to follow. Specifically for the GLXP mission, the global path plan has the added benefit of aiding the distance verification task. That is, the path can help verify the rover indeed travelled at least 550m on the lunar surface.

Future work may include online search methods. Feature recognition will likely miss obstacles, and smaller obstacles may still impede the rover's path. This uncertainty in the environment can be handled with online replanning [17]. As discussed previously, the D* search method is better suited for this task. D*, similar to A*, maintains an open list of nodes yet to be evaluated. The key difference is D* searches backwards from the goal node, whereas A* searches from start to goal. Each node expanded from the open list has a backpointer, referring to the next node leading to the goal, and each backpointer node knows the exact cost to the goal. The algorithm terminates when the start node is next to be expanded, with the solution path the sequence of backpointers. The *dynamic* difference from A* is in obstacle handling. When an obstacle is detected along the path, all affected nodes are placed on the open list, with a flag to raise the cost. Backpointers are checked to see if they should be raised as well. Similarly, a flag to lower the cost propagates through backpointers [20].

Path cost can also include other parameters such as terrain elevation and solar incidence. For the former, cost can vary in the freespace depending if the rover is traversing uphill, downhill, or sidehill. For the latter, solar incidence angles can be incorporated in order to favor direct sunlight on the solar array, and possibly increase the path cost for direct sunlight on thermally sensitive components. Expanding the cost metric in this manner would call for a *cost map*: the cost map is the same size as the occupancy grid and the value of each element represents the cost of traversing the cell. In the method descried here, traversing each cell is a constant unit cost. The costs of mission tasks are integrated into the path planning scheme in the work done by WU Peng and JU Hehua [21], planning the rover path and task sequence synchronously.

Although designed for a specific lunar mission pursuing the GLXP, the framework is suitable for other planetary rover missions with aerial imagery; the optimization scheme discussed here can be applied to rover missions on other moons, Mars, and asteroids.

## Acknowledgment

The author would like to thank Professors Matt Eicholtz and David Wettergreen of Carnegie Mellon University, and Heather Jones of Astrobotic for their continued support and mentorship.